\documentclass[10pt,twocolumn,letterpaper]{article}

\usepackage{wacv}
\usepackage{times}
\usepackage{epsfig}
\usepackage{graphicx}
\usepackage{amsmath}
\usepackage{amssymb}

\wacvfinalcopy 

\def\wacvPaperID{****} 

\ifwacvfinal\fi
\begin{document}

\title{Accuracy Booster: Performance Boosting using Feature Map Re-calibration}

\author{
Pravendra Singh \hspace{2cm}Pratik Mazumder \hspace{2cm}Vinay P. Namboodiri\\
Department of Computer Science and Engineering, IIT Kanpur, India\\
{\tt\small \{psingh, pratikm, vinaypn\}@iitk.ac.in}
}
\maketitle
\ifwacvfinal\thispagestyle{empty}\fi

\begin{abstract}
Convolution Neural Networks (CNN) have been extremely successful in solving intensive computer vision tasks. The convolutional filters used in CNNs have played a major role in this success, by extracting useful features from the inputs. Recently researchers have tried to boost the performance of CNNs by re-calibrating the feature maps produced by these filters, e.g., Squeeze-and-Excitation Networks (SENets). These approaches have achieved better performance by \textit{Exciting} up the important channels or feature maps while diminishing the rest. However, in the process, architectural complexity has increased. We propose an architectural block that introduces much lower complexity than the existing methods of CNN performance boosting while performing significantly better than them. We carry out experiments on the CIFAR, ImageNet and MS-COCO datasets, and show that the proposed block can challenge the state-of-the-art results. Our method boosts the ResNet-50 architecture to perform comparably to the ResNet-152 architecture, which is a three times deeper network, on classification. We also show experimentally that our method is not limited to classification but also generalizes well to other tasks such as object detection.
\end{abstract}

\section{Introduction}

Convolutional neural networks (CNNs) have surpassed many traditional machine learning approaches in solving several computer vision tasks such as classification \cite{krizhevsky2012imagenet,simonyan2014very}, segmentation \cite{chen2018deeplab}, detection \cite{ren2015faster,liu2016ssd} and others. Various works \cite{singh2019hetconvijcv,singh2019hetconv,singh2019stability,singh2018leveraging,singh2019multi,singh2019falf,mazumder2019cpwc,singh2019play,singh2020cooperative} have been proposed for efficient deep learning. Researchers have recently been trying to improve CNN performance, by promoting channels (feature maps) that are more relevant \cite{hu2017squeeze}. Each channel or feature map is produced by a Convolutional filter, and each Convolutional layer can have multiple such filters. Therefore, the significance of a feature map points to the relevance of the Convolutional filter that produced it. It has been experimentally shown that increasing the contribution of relevant channels towards creating higher-level features improves the performance of CNNs \cite{hu2017squeeze}. Therefore, the recent works have focussed on \textit{learning} the significance (re-calibration weights) of the feature maps. We will use the terms feature map and channel interchangeably to refer to the output produced by a Convolutional filter.

The \cite{hu2017squeeze} paper (SENet) captures this channel relevance and shows improvement over the base CNN models. The \cite{woo2018cbam} (CBAM) paper improves upon the SENet idea by using a combination of channel importance and spatial importance to learn better features and improve the network performance further. 

\begin{figure}[t]
    \centering
    \includegraphics[scale = 0.25]{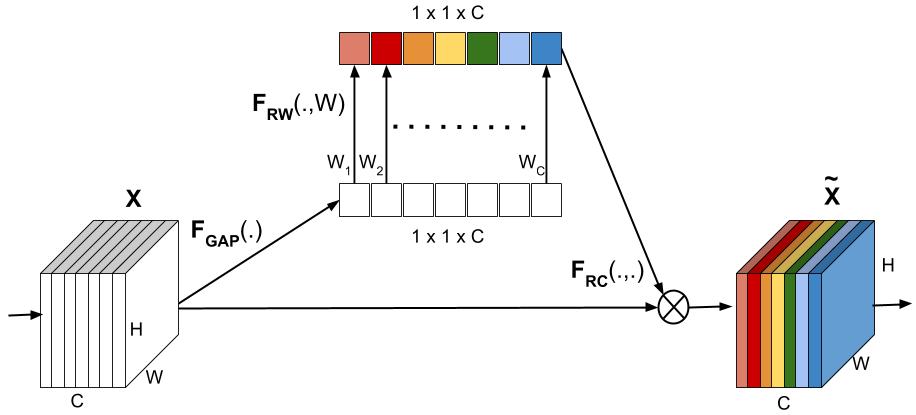}
    \caption{Figure shows the Accuracy Booster block with Depth Wise $1\times1$ Convolution (best viewed in color).}
    \vskip -0.2in
    \label{fig:AB}
   
\end{figure}

However, we find that these methods perform certain redundant transformations to find the re-calibration weights of feature maps, and in the process, they shoot up the architectural complexity of the base model. We propose an idea that performs better than these methods while requiring simpler and lighter modifications to the base networks as compared to these methods. Our method finds a single representative data-point for each channel and applies $1\times1$ Depth Wise Convolution operation with Batch Normalization (BN) and sigmoid activation to find the channel significance (re-calibration weights). Our lighter architecture, Accuracy Booster block (Figure \ref{fig:AB}), performs better than the existing methods. We also propose a heavier model (AB-Plus) that significantly beats all the other methods in performance while having a similar computation complexity (FLOPS).

In this work, we show that our design beats the state-of-the-art results in classification on ImageNet and CIFAR datasets and generalizes well for object detection on MS-COCO dataset. We analyze the effects of further reducing our model complexity in our detailed ablation studies.

Our major contributions are as follows:

\begin{itemize}
    \item We propose a simpler and more efficient Accuracy Booster block (AB) that significantly boosts the performance of CNNs. We provide experimental results that vindicate our choice of architecture through extensive ablation studies. 
    \item We show that our proposed block works well for various networks not only for classification but also for detection.
    \item We empirically show that our proposed block consistently performs better than SE blocks and other follow up architectures while introducing much lower complexity (extra FLOPs, Parameters, Runtime Memory) to the original model as compared the other blocks.
\end{itemize}

\section{Related Works}
\label{Related Works}

\textbf{Network Design Improvement:} Improving the network architecture of CNNs has remained a hot topic from some time now \cite{huang2017densely,zoph2018learning}. Improvements have targeted design changes that lead to better performance by the network on various tasks. With an increase in computation power and dataset size, researchers have looked towards even deeper network architectures for CNNs to improve their performance. Architectures like Inception models \cite{szegedy2015going} and VGGNet \cite{simonyan2014very} showed that increasing the depth of a network could significantly increase the quality of representations that it was capable of learning. While Deeper Architectures improved the performance of CNNs, they also introduced problems like vanishing gradients, longer training time and higher space requirements for training and deployment. ResNet \cite{he2016deep} proposed skip-connections based on identity mapping, which reduced the optimization issues of deep networks. This allowed for using deeper and more complex networks. WideResNet \cite{zagoruyko2016wide} restricted the network depth and used wider layers to improve the performance, thereby modifying this idea. ResNeXt \cite{xie2017aggregated} proposed parallel aggregated transformations blocks and showed that increasing the number of such parallel blocks led to better performance. Our proposed approach improves network performance without significantly increasing the network depth or complexity.

\textbf{Attention:} 
Attention mechanisms in a network give higher importance to the most relevant components of the information flowing through the network \cite{itti1998model,larochelle2010learning,mnih2014recurrent,chen2017sca,vaswani2017attention}.

Channel-wise attention provides the re-calibration weights for the channels (feature maps) generated by convolutional filters. The SE block \cite{hu2017squeeze} \textit{Squeezes} the output channels, finds the re-calibration weights for each channel and then \textit{Excites} the channels using these weights. But it also adds to the depth of the network, thereby increasing the complexity of the network and the network latency.  CBAM \cite{woo2018cbam} makes use of a combination of channel-wise attention and spatial attention to learn better representations for achieving the same goal. GE-$\Theta^{+}$ model \cite{hu2018gather} uses the excitation module of the SE block as a black box and experiments with the squeeze module.

Our proposed AB block uses a very simple transformation to get the re-calibration weights (RW) and still gets better performance than all the other existing methods derived from SE \cite{hu2017squeeze}. AB-Plus block achieves even better performance than the AB block but introduces a higher number of parameters with similar computational complexity.

All recent block architectures that have built upon the SE block architecture have increased the complexity of the base model further. Our Proposed AB block is the first to introduce lower complexity than the SE block and still perform consistently better than all other blocks.

\begin{figure*}[!htb]
    \centering
    \includegraphics[scale=0.25]{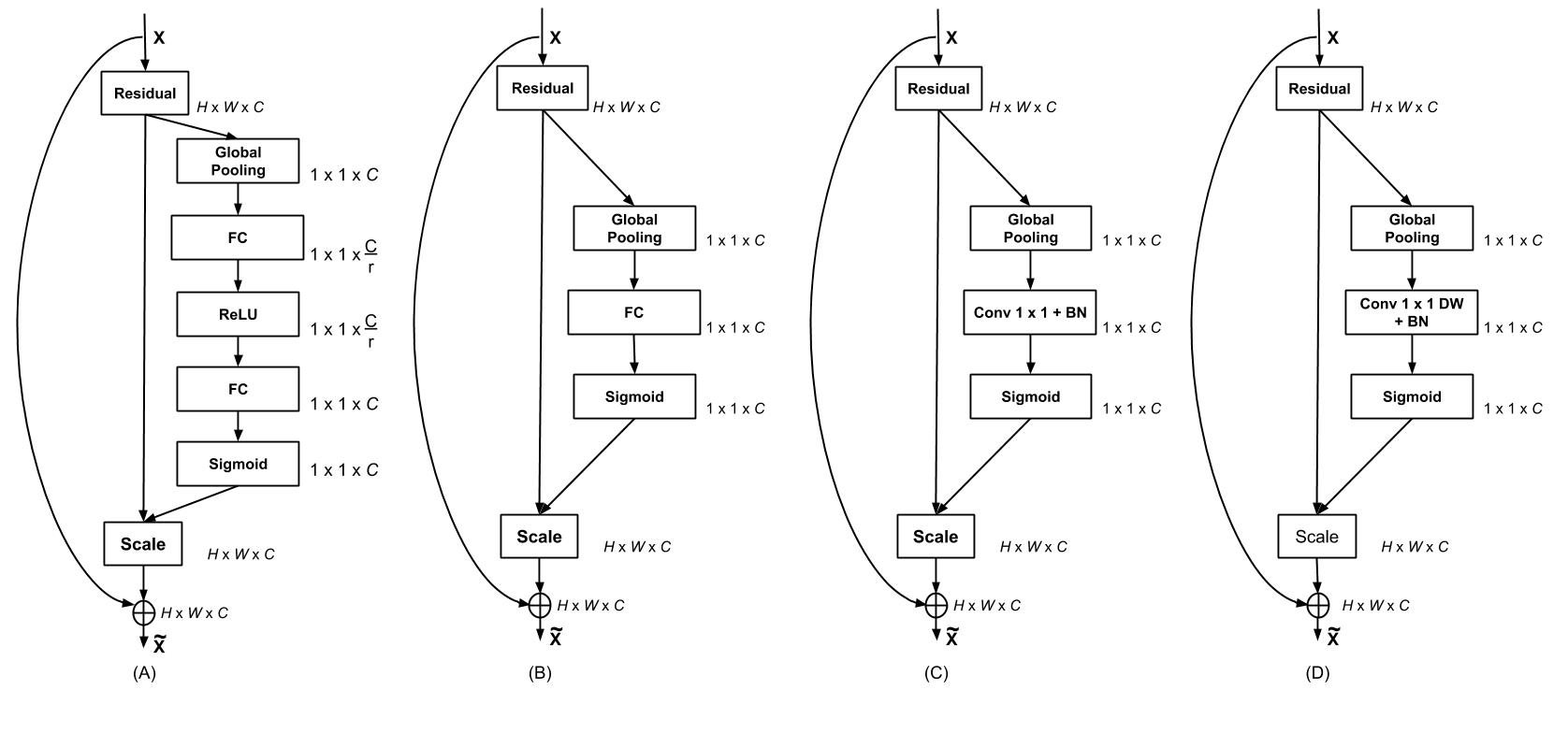}
    \caption{Figure shows the evolution of AB block from the SE block architecture. (A) SE block, (C) AB-Plus block, and (D) AB block. (B) and (C) are similar as fully  connected layer performs the same operation as Conv 1$\times$1 layer when operating on a 1$\times$1 feature maps.}
     \vskip -0.2in
    \label{fig:Evolution_SENet}
   
\end{figure*}

\section{Accuracy Booster Blocks}
\label{Accuracy Booster Blocks}
We propose two types of AB block architectures: AB (Fig \ref{fig:AB}) and AB-Plus (Fig \ref{fig:Evolution_SENet}(C)).

In the AB block, first, we need a representative for each channel (feature map), on the basis of which we can judge their importance. We use the Global Average Pooling ($F_{GAP}$) operator for this purpose. Our input feature maps $X\in \mathbf{R}^{H \times W \times C}$ consist of $C$ channels of height $H$ and width $W$. Therefore, given an input $X$, we compute a channel-wise representative for it, $Y \in \mathbf{R}^{C}$, using global average pooling as follows:
$$ y_k = F_{GAP}(x_k) = \frac{\sum_{i=1}^{H}\sum_{j=1}^{W} x_k(i,j)}{H \times W} $$
where each $y_k \in \mathbf{R}$ is a representative for the $k^{th}$ channel and $Y = [y_1,y_2,..,y_k,..,y_C]$ is the channel-wise descriptor. $x_k \in \mathbf{R}^{H \times W}$ is the feature map for $k^{th}$ channel of $X$ where $X = [x_1,x_2,..,x_k,..,x_C]$. $F_{GAP}$ is the the Global Average Pooling operator.

We use this simple technique (GAP) to get a descriptor for each channel because it adds no extra parameters. Other techniques for this purpose are discussed in the ablation studies.

Next, we need to use these representatives to find out the significance of each channel over the other ($F_{RW}$). We use $C$ Depth Wise $1\times1$ Convolution operators $W = [W_1,W_2,...,W_C]$, one for each of the channel representatives, where each $W_k \in \mathbf{R}$. This is followed by Batch Normalization (BN) and a sigmoid activation operator to get the re-calibration weight for each channel (Fig \ref{fig:AB}, Fig \ref{fig:Evolution_SENet}(D)). Formally, the re-calibration weights can be defined as,
$$P =  F_{RW}(Y) = \sigma(BN(DWConv(Y)))$$
Where $\sigma$ is the sigmoid activation operator, $P$ is the channel-wise re-calibration weights, DWConv is the Depth Wise $1\times1$ Convolution operator, BN is Batch Normalization applied after performing the Depth Wise Convolution operations. The DWConv can also be seen as the channel-wise product of the re-calibration weights in $W$ and the channel representatives in $Y$, i.e., $[W_1.y_1,W_2.y_2,...,W_C.y_C]$.

Finally, the input $X=[x_1,x_2,..,x_k,..,x_C]$ is re-calibrated using the learned re-calibration weights $P=[p_1,p_2,..,p_k,..,p_C]$ ie. $\widetilde{X} = F_{RC}(P,X) $, such that,
$$ \widetilde{x_k} = p_k.x_k $$
where, $p_k.x_k$ is a product of the scalar re-calibration weight $p_k$ and the feature $x_k \in \mathbf{R}^{H \times W}$ of $k^{th}$ the channel of the input. $\widetilde{X}=[\widetilde{x_1},\widetilde{x_2},..,\widetilde{x_k},..,\widetilde{x_C}] $ is the re-calibrated output feature maps.

The AB-Plus block is almost the same as the AB block, but it uses $C$ $1\times1$ Convolutional Operators instead of $C$ Depth Wise $1\times1$ Convolutional Operators. The difference is that each of the $C$ Depth Wise $1\times1$ Convolutiona1 layers learns $1$ scalar weight $W_k$ for mapping the $k_{th}$ channel descriptor $y_k$ to its re-calibration weight $p_k$,i.e. it can be thought of as a one to one connection between $2$ sets of $C$ nodes each. Whereas, each of the $C$ Full $1\times1$ Convolutional layers learns a weight vector $\widetilde{W_k} \in \mathbf{R}^{C}$ for using the entire channel-wise descriptor $Y$ to learn the re-calibration weight $p_k$ of the $k^{th}$ channel,i.e. it can be thought of as a fully connected layer between $2$ sets of $C$ nodes each. This means that the AB-Plus block has more parameters than the AB block.

The proposed block can be added after any convolutional layer. In the case of residual blocks, the proposed block should be added before the summation operator for the skip connection, as shown in Fig \ref{fig:Evolution_SENet}.

\section{Relation to SE-Net}
\label{Relation to SE-Net}

As mentioned earlier, the SE-Net \cite{hu2017squeeze} also learns channel (feature map) importance and uses them to promote the useful channels. As can be seen in Fig.\ref{fig:Evolution_SENet}(A), the SE block \textit{Squeezes} the channels produced by a Convolutional layer using Global Average Pooling, to get a channel-wise descriptor $Y = [y_1,y_2,..,y_k,..,y_C]$ where $y_k \in \mathbf{R}$ and $Y \in \mathbf{R}^{C}$. Next, a fully connected layer transforms $Y$ of size $C$ to another smaller descriptor of size $\frac{C}{r}$, where $r$ is a hyperparameter. Then, another fully connected layer transforms the smaller descriptor back to the original $C$ sized descriptor which is followed by a sigmoid activation operator to get the channel-wise re-calibration weights. These weights are used to \textit{Excite} the channels.

A shallower design (Fig.\ref{fig:Evolution_SENet}(B)) can have only one fully connected layer after the Global Average Pooling operator, which transforms the channel-wise descriptor $Y$ to another descriptor of the same size. The sigmoid activation is then applied to it, to get the re-calibration weights.

This design is equivalent to using  $C$ $1\times1$ Convolutional operators, which in effect is the same as a fully connected layer of $C$ nodes. This is our AB-Plus model (Fig.\ref{fig:Evolution_SENet}(C)).

Our proposed design (Fig.\ref{fig:Evolution_SENet}(D)) replaces the $C$ Convolutional operator in the previous with $C$ Depth Wise Convolutional operators. Depth Wise Convolutional operators create a one to one connection between the channel-wise descriptors, before and after the transformation, as can be seen in Fig.\ref{fig:AB}. This means our design is lighter than the model in Fig.\ref{fig:Evolution_SENet}(C) and significantly lighter than the SE-Block Fig.\ref{fig:Evolution_SENet}(A). 

To the best of our knowledge, this is the first work that reduces the complexity of the SE block while consistently performing better than it.

Our Experiments and Ablation studies show that the compressing and expanding of the channel-wise descriptor by the SE block prevents the SE block from achieving the full potential of improvements that can be obtained by using channel-wise re-calibrations. Our approach improves the CNN performance further by overcoming this architectural drawback.

Further, complexity analysis of our proposed design and previous designs are given in the next section.

\section{Analysis of Model Complexity}
\label{Analysis of Model Complexity}
One of the major goals of Network Architecture improvement is to avoid increasing the network complexity significantly in the process of improving network performance. We compare the complexity introduced by our block to that done by the other recent designs. We compare the designs on the basis of the extra parameters, extra computation, and extra run time memory requirement introduced by the performance-boosting approaches. 
\begin{table}[t]
  \begin{center}
    \caption{Analysis of Extra Parameters, Extra FLOPS, and  Extra Run Time Memory (RTM) introduced per block for B batch size.}
    \label{tab:extra}
    \scalebox{.95}{
    \renewcommand{\arraystretch}{1.5}
     \begin{tabular}{|l|c|c|c|} 
      \hline
     \textbf{Models} & AB & SE & CBAM\\
      \hline
      \textbf{Extra Params.} & $C$   & $\frac{2C^2}{r}$ &  $>\frac{2C^2}{r}$\\ 
      \hline
      \textbf{Extra FLOPS} & $C.B$   & $\frac{2C^2.B}{r}$ & $>\frac{2C^2.B}{r}$\\
      \hline
      \textbf{RTM} & $4C(1+B)$   & $ \frac{8C(C+B)}{r} $ & $ >\frac{8C(C+B)}{r} $\\
     \hline
    \end{tabular}}
  \end{center}
  \vskip -0.3in
\end{table}

\subsection{Extra Parameters}
\label{Extra Parameters}
 As mentioned in Table \ref{tab:extra}, the AB block introduces only $C$ extra parameters which are the parameters in the $C$ Depth Wise $1\times1$ Convolutional operators. The SE block adds $\frac{2C^2}{r}$ since it uses $2$ fully connected layers of size $\frac{C}{r}$ and $C$ respectively. The CBAM block uses the SE block for the channel attention along with another subnetwork for spatial attention. Therefore, the number amount of extra parameters it introduces is greater than that of SE. So we can see that the AB block introduces only extra parameters of the order of $O(C)$, which very less compared to the others, and still performs better. The AB-Plus block adds $C^2$ extra parameters, which makes it heavier than the others, but it performs significantly better than all the other designs.

\subsection{Extra FLOPS}
\label{flops}

FLoating point OPerations per Second (FLOPS) for a model can be used to describe its computational complexity. FLOPS can be used to represent the total number of computations. The FLOPS are calculated using the process described in \cite{singh2019stability}.

As mentioned in Table \ref{tab:extra}, the AB block introduces only $C.B$ extra FLOPS. The SE block adds $\frac{2C^2.B}{r}$ extra FLOPS. Since the CBAM block uses the SE block for the channel attention along with another subnetwork for spatial attention, therefore it introduces extra FLOPS greater than that of SE. So the AB block requires lesser extra FLOPS than both of them while the AB-Plus model adds $C^2.B$ extra FLOPS.

\subsection{Extra Run Time Memory Size Requirements}
\label{RTM}

Run Time Memory (RTM) denotes the memory space required to store the feature maps and the model parameters. The extra Run Time Memory is calculated using the process described in \cite{singh2019stability}.

As mentioned in Table \ref{tab:extra}, the AB block introduces only $C \times 4+1 \times 1 \times C \times 4 \times B= 4C(1+B)$ extra RTM. The SE block adds $\frac{2C^2}{r}\times 4 + \frac{2C}{r}\times 4 \times B =  \frac{8C(C+B)}{r}$ extra RTM.  Since the CBAM block uses the SE block for the channel attention along with another subnetwork for spatial attention, therefore it introduces extra RTM greater than that of SE. So the AB block has the lowest extra RTM requirement.

\section{Experiments}
\label{experiments}
This section explores the experimental results of incorporating the AB block into various CNN architectures for various tasks and datasets.

\subsection{Image Classification}
\label{classification}
In this section, we explore how the AB block improves the performance of networks like ResNet-50 and others on the image classification task.

Experiments are conducted on the ImageNet large scale dataset \cite{russakovsky2015imagenet}, CIFAR-10 and CIFAR-100 datasets \cite{krizhevsky2009learning}. The ImageNet dataset has around 1.28 million training images and 50000 validation images from 1000 different classes. The training is done on the training set, and the Top-1 and Top-5 errors on the validation set are reported.

For the ImageNet dataset, we verify the performance of ResNet-50 \cite{he2016deep}, WideResNet-18 (widen=2) \cite{zagoruyko2016wide} with and without the AB block.
For the CIFAR-10 and CIFAR-100 datasets, we verify the performance of ResNet-56 \cite{he2016deep}, ResNet-164 \cite{he2016deep}, WideResNet-22 (widen=10) \cite{zagoruyko2016wide} with and without the AB block.
For ImageNet experiments, we use the same settings and setup as mentioned in CBAM \cite{woo2018cbam}.
The results on the CIFAR-10/100 datasets for all the architectures have been reproduced in the PyTorch  \cite{paszke2017automatic}.

We also compare our AB and AB-Plus results with the networks modified with SE blocks and CBAM blocks. The additional tricks used by SE Blocks, such as repeated training with lower learning rate when the loss plateaus and additional augmentation techniques such as pixel jittering, image rotation, are not used in our experiments in order to maintain standard conditions. 

\begin{table}[t]
  \begin{center}
    \caption{Single-crop error rate ($\%$) on the ImageNet validation set and complexity comparisons for ResNet-50.}
    \label{tab:imagenetresnet}
    \scalebox{.8}{
    
     \begin{tabular}{|l|c|c|c|c|} 
      \hline
    \textbf{Models} & \textbf{Top-1} & \textbf{Top-5} & \textbf{Params} & \textbf{FLOPS} \\
    \hline
    ResNet-50 (Baseline) \cite{woo2018cbam,he2016deep} & $24.56$ & $7.50$ & $25.56$M & $3.858$G \\
    \hline
    ResNet-152 \cite{hu2017squeeze,he2016deep}  & $22.42$ & $6.34$ & $60.19$M & $11.30$G \\
    \hline
        SE  \cite{woo2018cbam,hu2017squeeze}          & $23.14$ & $6.70$ & $28.09$M & $3.860$G \\
    \hline
    CBAM   \cite{woo2018cbam}         & $22.66$ & $6.31$ & $28.09$M & $3.864$G \\
    \hline
    AB  (Ours)          & $\textbf{22.4}$ & $\textbf{6.2}$ & $25.57$M & $3.858$G \\
    \hline
    AB-Plus  (Ours)          & $\textbf{22.1}$ & $\textbf{6.1}$ & $45.67$M & $3.878$G  \\
      \hline
    \end{tabular}}
  \end{center}
  \vskip -0.2in
\end{table}
As can be seen in Table \ref{tab:imagenetresnet}, AB blocks significantly improve the network performance over the baseline ResNet-50 and also exceed the improvement produced by the SE and CBAM blocks. The ResNet-50 network with AB blocks also performs comparably to the ResNet-152 network which is $3$ times deeper, uses more than twice the number of parameters and requires almost $3$ times the FLOPS required by the ResNet-50 network with AB blocks. We can also see that the SE and CBAM blocks add $2.5$M more parameters to base ResNet-50 network, whereas our AB blocks just add $0.01$M extra parameters. AB-Plus blocks further improve the performance, but since it introduces more parameters, there is a trade-off. Though the AB-Plus blocks introduce a lot of parameters, the FLOPS do not increase by much because the extra parameters belong only to the $1\times1$ Convolutions applied on $1\times1\times C$ sized feature maps. 

\begin{table}[t]
  \begin{center}
    \caption{Inference Time per batch needed for base ResNet-50 network and base modified by SE and AB blocks on ImageNet dataset.}
    \label{tab:timecmp}
    \scalebox{.85}{
     \begin{tabular}{|l|c|c|c|} 
      \hline
     \textbf{Models} & \textbf{Batch Size}  & \textbf{ Time in sec} & \textbf{FLOPS}\\ 
     \hline
      ResNet-50 (Baseline) &  $256$  & $0.152$ & $3.858$G\\
     \hline
      SE &  $256$  & $0.187$ & $3.860$G\\ 
     \hline
      AB (Ours) & $256$  & $0.163$ & $3.858$G\\
     \hline
     AB-Plus (Ours) & $256$  & $0.192$ & $3.878$G\\
     \hline
    \end{tabular}}
  \end{center}
  \vskip -0.2in
\end{table}

\begin{table}[t]
  \begin{center}
    \caption{Single-crop error (Top-1 and Top-5 error rate) ($\%$)  on the ImageNet validation set and complexity comparisons for WideResNet-18 with widen=2 (WRN).}
    \label{tab:imagenetwrn}
    \scalebox{.79}{
     \begin{tabular}{|l|c|c|c|c|} 
      \hline
    \textbf{Models} & \textbf{Top-1} & \textbf{Top-5} & \textbf{Params} & \textbf{FLOPS} \\   
    \hline
    WRN-18-2 (Baseline) \cite{woo2018cbam,zagoruyko2016wide} & $25.63$ & $8.20$ & $45.62$M & $6.696$G \\
    \hline
        SE   \cite{woo2018cbam,hu2017squeeze}         & $24.93$ & $7.65$ & $45.97$M & $6.696$G \\
    \hline
    CBAM   \cite{woo2018cbam}         & $24.84$ & $7.63$ & $45.97$M & $6.697$G  \\
    \hline
    AB  (Ours)          & $\textbf{24.7}$& $\textbf{7.6}$ & $45.62$M & $6.696$G \\
    \hline
    AB-Plus  (Ours)          & $\textbf{24.6}$ & $\textbf{7.5}$ & $48.40$M & $6.698$G  \\
      \hline
    \end{tabular}}
  \end{center}
  \vskip -0.2in
\end{table}
From Table \ref{tab:imagenetwrn}, we can also see that AB blocks significantly improve the network performance over the baseline WideResNet-18 (widen=2) and those modified by the SE and CBAM blocks. AB-Plus blocks further reduce the error rates for WideResNet.

\begin{table}[t]
  \begin{center}
    \caption{Classification error ($\%$) on the CIFAR-10 for ResNet-56, ResNet-164, WideResNet-22-10.}
    \label{tab:cifar10}
    \scalebox{.85}{
    \addtolength{\tabcolsep}{3pt}
     \begin{tabular}{|l|c|c|c|c|} 
      \hline
     \textbf{Models} & \textbf{Original} & \textbf{SE} & \textbf{AB} & \textbf{AB-Plus}\\ 
     \hline
      ResNet-56 &  $6.5$ & $5.6$ & $\textbf{5.34}$& $\textbf{5.26}$ \\
     \hline
      ResNet-164 & $5.5$ & $4.4$ & $\textbf{4.0}$& $\textbf{3.9}$ \\
     \hline
      WideResNet-22-10 & $4.4$ & $4.0$ & $\textbf{3.7}$& $\textbf{3.7}$ \\
     \hline
    \end{tabular}}
  \end{center}
  \vskip -0.3in
\end{table}

From the Table \ref{tab:cifar10}, we can see that the networks modified with AB blocks show significant improvement in performance over the baseline ResNet-56 and ResNet-56 modified with the SE blocks for classification over the CIFAR-10 dataset. The AB modified ResNet-56 even performs similarly to the deeper ResNet-164 base model. AB-Plus blocks further reduce the error rate. 

\begin{table}[t]
  \begin{center}
    \caption{Classification error ($\%$) on the CIFAR-100 for ResNet-56, ResNet-164, WideResNet-22 (widen=10).}
    \label{tab:cifar100}
    \scalebox{.85}{
    \addtolength{\tabcolsep}{2pt}
     \begin{tabular}{|l|c|c|c|c|} 
      \hline
     \textbf{Models} & \textbf{Original} & \textbf{SE} & \textbf{AB} &\textbf{ AB-Plus}\\ 
     \hline
      ResNet-56 &  $28.6$ & $27.3$  & $\textbf{26.9}$& $\textbf{26.3}$ \\
     \hline
      ResNet-164 & $24.3$ & $21.8$  & $\textbf{21.5}$& $\textbf{21.3}$ \\
     \hline
      WideResNet-22-10 & $20.6$ & $19.3$  & $\textbf{19.0}$& $\textbf{18.8}$ \\
     \hline
    \end{tabular}}
  \end{center}
  \vskip -0.2in
\end{table}

From the Table \ref{tab:cifar100}, we can see that the networks modified with AB blocks show significant improvement in performance over the baseline ResNet-56 dataset and ResNet-56 modified with the SE blocks for classification over the CIFAR-100 dataset. AB-Plus blocks further reduce the error rates. If we remove Batch Normalization from the AB block, we observe 0.3\%, 0.2\%, and 0.2\% reduction in the accuracy reported in Table \ref{tab:cifar100} for ResNet-56, ResNet-164, and WideResNet-22-10 respectively.

\subsubsection{Inference Time}
From Table \ref{tab:timecmp}, we can see that our modification results in a significant reduction in the inference time of the SE modified ResNet-50 on ImageNet dataset. The AB model reduces the extra inference time needed by the SE block by one-third because the SE block introduces more latency by using $2$ layers in the calibration process while the AB block uses $1$ layer. Therefore, the AB block has a lower inference time than the SE block, although the FLOPS for the models are similar. All these experiments were run on a single Titan X GPU, with batch size $256$. Since CBAM \cite{woo2018cbam} and GE-$\Theta^{+}$ \cite{hu2018gather} both use the SE block as a black box and adds further layers/computation on top of it, they are bound to have more inference time that the SE block and the AB block.

\subsubsection{Mobile-optimized networks} We verify the performance of the AB blocks on mobile-optimized networks such as ShuffleNet \cite{zhang2017shufflenet}. We use the same training settings, as mentioned in the SE-Net paper \cite{hu2017squeeze}. Apart from comparing our results with the baseline ShuffleNet, we also compare our results with the ShuffleNet with SE blocks. As can be seen in Table \ref{tab:shuffle}, AB and AB-Plus blocks significantly improve the network performance over the baseline network and also exceed the improvement produced by the SE blocks.

\begin{table}[t]
  \begin{center}
    \caption{Performance of Mobile optimised networks: Single-crop error rates ($\%$) on the ImageNet validation set and complexity comparisons for ShuffleNet.}
    \label{tab:shuffle}
    \scalebox{.8}{
     \begin{tabular}{|l|c|c|c|c|} 
      \hline
     \textbf{Models} & \textbf{Top-1}& \textbf{Top-5}& \textbf{Params}& \textbf{FLOPS}\\ 
      \hline
      ShuffleNet (Baseline) \cite{hu2017squeeze,zhang2017shufflenet}  &  $32.6$ & $12.5$ & $1.80$M & $140.0$M \\
     \hline
      SE \cite{hu2017squeeze,zhang2017shufflenet} & $31.0$ & $11.1$ & $2.40$M & $142.0$M \\
     \hline
     \textbf{AB (Ours)} & $\textbf{30.5}$ & $\textbf{11.0}$ & $1.82$M & $140.5$M\\
     \hline
      \textbf{AB-Plus (Ours)} & $\textbf{30.3}$ & $\textbf{10.9}$ & $6.70$M & $145.3$M\\
     \hline
    \end{tabular}}
  \end{center}
  \vskip -0.3in
\end{table}

\begin{table}[t]
  \begin{center}
    \caption{Single-crop error rate ($\%$) on the ImageNet validation set and complexity comparisons for ResNeXt-101 (32$\times$4d).}
    \label{tab:imagenetresnext}
    \scalebox{.8}{
    
     \begin{tabular}{|l|c|c|c|c|} 
      \hline
    \textbf{Models} & \textbf{Top-1} & \textbf{Top-5} & \textbf{Params} & \textbf{FLOPS} \\
    \hline\hline
    ResNeXt-101 (Baseline)  & $21.2$ & $5.6$ & $44.18$M & $7.99$G \\\hline
    SE            & $20.7$ & $5.01$ & $48.96$M & $8.00$G \\
    \hline
    \hline
    GE-$\theta^{+}$          & $20.5$ & $4.8$ & $57.92$M & $8.02$G \\
    \hline
    \textbf{GE-$\theta^{+}$ with AB (Ours)} & $\textbf{20.2}$ & $\textbf{4.6}$ & $\textbf{53.23}$M & $\textbf{8.00}$G \\\hline
    \end{tabular}}
  \end{center}
   \vskip -0.2in
\end{table}

\subsubsection{Improving GE-$\Theta^{+}$ using AB block}
The top-performing model (GE-$\theta^{+}$) of \cite{hu2018gather} uses the excitation phase of SE (2 fully connected layers with reduction ratio $r=16$) as a black box to perform excitation. Therefore, if we replace the SE Excitation module used in GE-$\theta^{+}$ by our proposed Excitation module ($1\times1$ DW Conv + BN), then the performance improvement is guaranteed since our proposed excitation module performs significantly better than the SE excitation module. 

In Table~\ref{tab:imagenetresnext}, GE-$\theta^{+}$ with AB is the model which uses our proposed Excitation module ($1\times1$ DW Conv + BN) in place of SE Excitation module (two FC with $r=16$) in GE-$\theta^{+}$. From Table~\ref{tab:imagenetresnext} we can see that GE-$\theta^{+}$ with AB significantly reduces the number of parameters (by around 4.7 Million) when compared to GE-$\theta^{+}$ and also performs better than it.

\subsection{Object Detection}
\label{detection}
\begin{table}[t]
  \begin{center}
    \caption{Performance on Object Detection: Object detection mAP ($\%$) on the MS COCO validation set using Faster R-CNN.}
    \label{tab:detection}
    \scalebox{.85}{
    \addtolength{\tabcolsep}{2pt}
     \begin{tabular}{|l|c|c|} 
      \hline
     \textbf{Base Model} & \textbf{AP@IoU=0.5} & \textbf{AP@IoU=0.5:0.95}\\ 
      \hline
      ResNet-50 \cite{hu2017squeeze,ren2015faster} &  $45.2$ & $25.1$ \\
     \hline
      SE ResNet-50 \cite{hu2017squeeze}&  $46.8$ & $26.4$ \\
     \hline
     AB ResNet-50 (Ours) &  $\textbf{47.2}$ & $\textbf{26.7}$ \\
     \hline
    \end{tabular}}
  \end{center}
  \vskip -0.3in
\end{table}

We explore how the AB block improves object detection performance. We use the MS-COCO dataset \cite{lin2014microsoft}. It consists of around 80,000 training and  40,000 validation images. The ResNet-50 model used in the Faster R-CNN network \cite{ren2015faster} is modified with AB blocks to explore how the AB blocks generalize well to object detection tasks.

Table \ref{tab:detection} shows the validation set performance of the object detector using the base ResNet-50 and the modified ResNet-50 with the AB and SE blocks. The AB modified ResNet-50 shows improvement over the base and SE model.

We can conclude from these experiments that AB blocks induce better improvements in the network performance across a number of architectures, datasets and task than other existing methods.

\section{Ablation Study}
\label{ablation}
We perform ablation experiments on the AB block architecture to explore the significance of the design choices that we have made. Performance of AB modified ResNet-50/ResNet-56 is computed for Classification task on ImageNet/CIFAR-100 dataset. The Standard data augmentation strategy of random crop and random horizontal flip is used for carrying out the ablation experiments. All the architectures have been extensively trained with the same settings for fair comparisons.

\subsection{Network Depth}
\label{depth}
\begin{table}[t]
  \begin{center}
    \caption{Classification accuracy $\%$ on CIFAR-100 dataset for the ResNet-56 network with modifications as given in Fig \ref{fig:Evolution_SENet} (A,C,D).}
    \label{tab:evolution}
    \scalebox{.85}{
    \addtolength{\tabcolsep}{6pt}
     \begin{tabular}{|l|c|c|} 
      \hline
     \textbf{Models} & \textbf{Accuracy($\%$)} & \textbf{Depth}\\ 
      \hline
      ResNet-56 (baseline) &  $71.4$ & $0$\\
     \hline
      SE (2 FC layers) & $72.7$ & $2$\\
     \hline
     \textbf{AB ($\textbf{$\textbf{1}\times\textbf{1}$}$ DWConv+BN)} & $\textbf{73.1}$& $1$\\
     \hline
     \textbf{AB-Plus ($\textbf{$\textbf{1}\times\textbf{1}$}$ Conv+BN)} & $\textbf{73.7}$& $1$\\
     \hline
    \end{tabular}}
  \end{center}
  \vskip -0.3in
\end{table}
As shown in Fig \ref{fig:Evolution_SENet}, our AB block (Fig.\ref{fig:Evolution_SENet}(D)) reduces the number of transformations used in the SE block.
The SE block (Fig.\ref{fig:Evolution_SENet}(A)) uses $2$ fully connected (FC) layers. A shallower design (Fig.\ref{fig:Evolution_SENet}(B)) can have only one FC layer after the Global Average Pooling, which transforms the channel-wise descriptor $Y$ to another descriptor of the same size.
This is equivalent to using  $C$ $1\times1$ Convolutional operators which is in effect the same as a fully connected layer of $C$ nodes (Fig.\ref{fig:Evolution_SENet}(C)). This is the architecture of our AB-Plus block. Our AB block (Fig.\ref{fig:Evolution_SENet}(D)) replaces the $C$ Convolutional operators in the previous design with $C$ Depth Wise Convolutional operators.

We checked the performance improvement induced by the designs in Fig \ref{fig:Evolution_SENet}(A,C,D) on the ResNet-56 network on Classification accuracy.

Table \ref{tab:evolution} shows that the AB design (Fig \ref{fig:Evolution_SENet}(D)) beats the SE design (Fig \ref{fig:Evolution_SENet}(A), $2$ FC layers). This shows that our choice of architecture (having fewer parameters and lower computational complexity than SE) does not hurt the CNN performance but improves it further. The AB-Plus design (Fig \ref{fig:Evolution_SENet}(C)) beats all the other designs but has more parameters (with similar computational complexity) than all the others. 

\subsection{Average Pooling}
\label{avg}

We further experimented with the AB block design by removing the $1\times1$ Depth Wise Convolutional layer. The first design (E) used only Global Average Pooling (GAP) to get the channel-wise descriptors followed by a sigmoid operation to get the re-calibration weights. The second design (F) modified the design (E) by using 1D Batch Normalization after the GAP operation. Since the GAP operation gives equal importance to all the points in each channel while finding the average, we checked if a weighted average with learnable weights can improve the performance further. The third design (G), uses $C$ Depth Wise Convolutional operators of the same spatial size as the channels or feature maps ($W\times H$) produced by the Convolutional filters of the base model. Each of the $C$ Depth Wise Convolutional operators is for one of the $C$ channels. So it learns a weight for each point in the input feature map. The fourth design (H), uses $C$ Convolutional operators of the same size as the set of feature maps ($W\times H \times C$) given as input to the block.

From Table \ref{tab:depthzero}, we can see that using ``Only GAP" (E design) improves the network performance over the baseline but still lags behind the performance of the AB blocks. Adding Batch Normalization to GAP (F design) is of no help too. The Depth Wise Global Weighted Average method (G design) improves the performance but still falls short of our AB design. The Global Weighted Average design (H) exceeds even the AB-Plus block results. But it is not practical to use this design since it adds $W\times H\times C\times C$ parameters for every block and will drastically shoot up the computational complexity (in the order of $O(WHC^2)$ per block.
\begin{table}[t]
  \begin{center}
    \caption{Classification accuracy $\%$ on CIFAR-100 dataset for the AB ResNet-56 network with modifications on the averaging process after removing the $1\times1$ DWConv layer in AB.}
    \label{tab:depthzero}
    \scalebox{.85}{
    \addtolength{\tabcolsep}{9pt}
     \begin{tabular}{|l|c|} 
      \hline
     \textbf{Models} & \textbf{Accuracy($\%$)} \\ 
      \hline
      ResNet-56 (baseline) &  $71.4$ \\
      \hline
      AB ($1\times1$ DWConv+BN) & $73.1$\\
     \hline
     AB-Plus ($1\times1$ Conv+BN) & $73.7$\\
     \hline
     \hline
      (E) Only GAP & $72.0$ \\
      \hline
      (F) Only GAP + BN & $71.9$ \\
      \hline
      (G) Global DW Wt. Avg & $72.8$ \\
     \hline
      (H) Global Wt. Avg & $\textbf{73.8}$ \\
     \hline
     
    \end{tabular}}
  \end{center}
  \vskip -0.2in
\end{table}

\subsection{Calibration Level}
\label{calibrationlevel}

\begin{table}[t]
  \begin{center}
    \caption{Classification accuracy $\%$ on CIFAR-100 dataset for the ResNet-56 network with modifications in AB block to use calibration for each point in the Feature Map (fine-grained calibration).}
    \label{tab:caliblevel}
    \scalebox{.85}{
    \addtolength{\tabcolsep}{12pt}
     \begin{tabular}{|l|c|} 
      \hline
     \textbf{Models} & \textbf{Accuracy($\%$)} \\ 
      \hline
      (I) $3\times3$ Conv & $71.36$ \\
      \hline
      (J) $3\times3$ DW Conv& $70.6$ \\
      \hline
      (K) $7\times7$ Conv & $72.3$ \\
     \hline
      (L) $7\times7$ DW Conv & $72.3$ \\
     \hline
     
    \end{tabular}}
  \end{center}
  \vskip -0.3in
\end{table}

In the AB block, we learn a single calibration weight for each channel. We experiment with the concept of learning calibration weights for each point in the feature map. The first design (I) uses $C$ $3\times3$ Convolutional operators with padding=1 to get another same sized point-wise descriptor for the set of feature maps ($\mathbf{R}^{W\times H\times C}$). This is then passed through a sigmoid operator, which gives separate re-calibration weights for each point in the set of channels (feature maps). The re-calibrated output is obtained by performing element-wise multiplication (Hadamard product) of each point in the original set of channels and in the re-calibration matrix. The second design (J) modifies the design (I) to use $C$ $3\times3$ Depth Wise Convolutional operators instead. The third (K) and fourth (L) design are similar to the design (I) and (J), except that they use $7\times7$ sized kernels for convolution with padding=3 so that a same sized (same size as the input set of channels $\mathbf{R}^{W\times H\times C}$) re-calibration matrix can be obtained. 

From Table \ref{tab:caliblevel}, we can see that the designs with $3\times3$ filters perform worse than the baseline. The $7\times7$ filter designs perform better but do not come close to the performance of AB blocks. 

\begin{table}[t]
  \begin{center}
    \caption{Classification accuracy $\%$ on ImageNet dataset for the ResNet-50 network with modifications in AB block to use calibration for each point in the Feature Map (fine-grained calibration).}
    \label{tab:caliblarge}
    \scalebox{.85}{
    \addtolength{\tabcolsep}{12pt}
     \begin{tabular}{|l|c|} 
      \hline
     \textbf{Models} & \textbf{Accuracy($\%$)} \\ 
     \hline
      ResNet-50 (baseline) &  $75.44$ \\
      \hline
      AB ($1\times1$ DWConv+BN) & $77.6$\\
     \hline
     AB-Plus ($1\times1$ Conv+BN) & $77.9$\\
     \hline
     \hline
      $3\times3$ DW Conv + BN& $76.6$ \\
      \hline
      $9\times9$ DW Conv + BN& $77.2$ \\
     \hline
     
    \end{tabular}}
  \end{center}
  \vskip -0.2in
\end{table}

We also perform fine-grained calibration (for each point) on the large scale dataset, Imagenet using the ResNet-50 architecture. We use $3\times3$ and $9\times9$ Depth Wise Convolution with suitable padding to get re-calibration weights for each point in the set of feature maps. The results in Table \ref{tab:caliblarge} show that both the designs are unable to beat the performance of the AB blocks.
Therefore, fine-grained calibration is unable to beat channel-wise calibration.

\subsection{Combining two types of Calibration Levels}
\label{calibrationlevelcombn}

We also experiment with using the two types of calibration (channel-wise and fine-grained) simultaneously. The first design (M) combines the design (L) ($7\times7$ DW Conv) with a GAP operator followed by a sigmoid to get a final channel-wise relevance. The second design (N) combines the design (L) ($7\times7$ DW Conv) with the AB block (GAP + $1\times1$ DWConv + BN) design. It consists of $C$ $7\times7$ Depth Wise convolutional operators with padding=3 to get another same sized point-wise descriptor for the set of feature maps ($\mathbf{R}^{W\times H\times C}$). This is followed by a GAP operator, $C$ $1\times1$ Depth Wise Convolutional operators, Batch Normalization and the sigmoid operator as present in the AB block. The third design (O) is same as the design (N) but uses the AB-Plus block (GAP + $1\times1$ Conv + BN) in place of the AB block design, which uses $1\times1$ Convolutional operators in place of Depth Wise Convolution.

\begin{table}[t]
  \begin{center}
    \caption{Classification accuracy $\%$ on CIFAR-100 dataset for the ResNet-56 network with modifications in AB block to use the $2$ types of calibration fine-grained and channel-wise.}
    \label{tab:caliblevelcomb}
    \scalebox{.85}{
     \begin{tabular}{|l|c|} 
      \hline
     \textbf{Models} & \textbf{Accuracy($\%$)} \\ 
      \hline
      ResNet-56 (baseline) &  $71.4$ \\
      \hline
      SE & $72.7$\\
      \hline
      AB ($1\times1$ DWConv+BN) & $73.1$\\
     \hline
     AB-Plus ($1\times1$ Conv+BN) & $73.7$\\
     \hline
     \hline
      (M) $7\times7$ DW Conv + GAP & $72.8$ \\
     \hline
      (N) $7\times7$ DW Conv + GAP + $1\times1$ DW Conv & $73.0$ \\
      \hline
      (O) $7\times7$ DW Conv + GAP + $1\times1$ Conv & $73.5$ \\
      \hline
    \end{tabular}}
  \end{center}
\vskip -0.3in
\end{table}
From Table \ref{tab:caliblevelcomb}, we can see that the design (N), which uses the AB block, performs worse than the standalone AB block. The design O, which uses the AB-Plus block, performs worse than the standalone AB-Plus block. Therefore, adding the fine-grained re-calibration to the channel-wise re-calibration results in better performance than the base network and the SE modified network but fails to reach the performance of the AB blocks. Also, since this approach combines two types of calibration, it has more parameters and computations than the AB blocks.

\subsection{Group Number}
\label{groupnumber}
We perform experiments on the number of groups (from 1 to number of channels) in the Convolution operation in the AB block. As can be seen in Table \ref{tab:cifargroup} there is no significant performance improvement by reducing the group number. 

\begin{table}[t]
  \begin{center}
   \caption{Classification Accuracy ($\%$) on the CIFAR-100 test set and complexity comparisons for ResNet-164.}
    \label{tab:cifargroup}
    \scalebox{.85}{
        \addtolength{\tabcolsep}{-2.5pt}
     \begin{tabular}{|l|c|c|c|} 
      \hline
    \textbf{Models} & \textbf{Acc(\%)} & \textbf{Params} & \textbf{FLOPS} \\
    \hline
    ResNet-164 (Baseline)  & 75.70  & $1.734$M & $246.58$M \\
    \hline
    \textbf{AB  G= \#Channel (DW conv)}   & \textbf{78.50}  & $\textbf{1.758}$M & $\textbf{246.59}$M \\
    \hline
    AB  G (number of groups) = 16   & 78.53 & $1.847$M & $246.68$M \\
    \hline
    AB  G (number of groups) = 8   & 78.57  & $1.944$M & $246.78$M \\
    \hline
    AB  G (number of groups) = 4   & 78.62  & $2.137$M & $246.97$M \\
    \hline
    AB  G (number of groups) = 2   & 78.66  & $2.524$M & $247.36$M \\
    \hline
    \textbf{AB-Plus G=1 (Standard conv)}   & \textbf{78.70}  & $3.298$M & $248.14$M  \\
      \hline
    \end{tabular}}
  \end{center}
  \vskip -0.2in
\end{table}

\subsection{Relevance of Calibration}
\label{releavancecalibration}

\begin{figure}[t]
    \centering
    \includegraphics[scale=0.45]{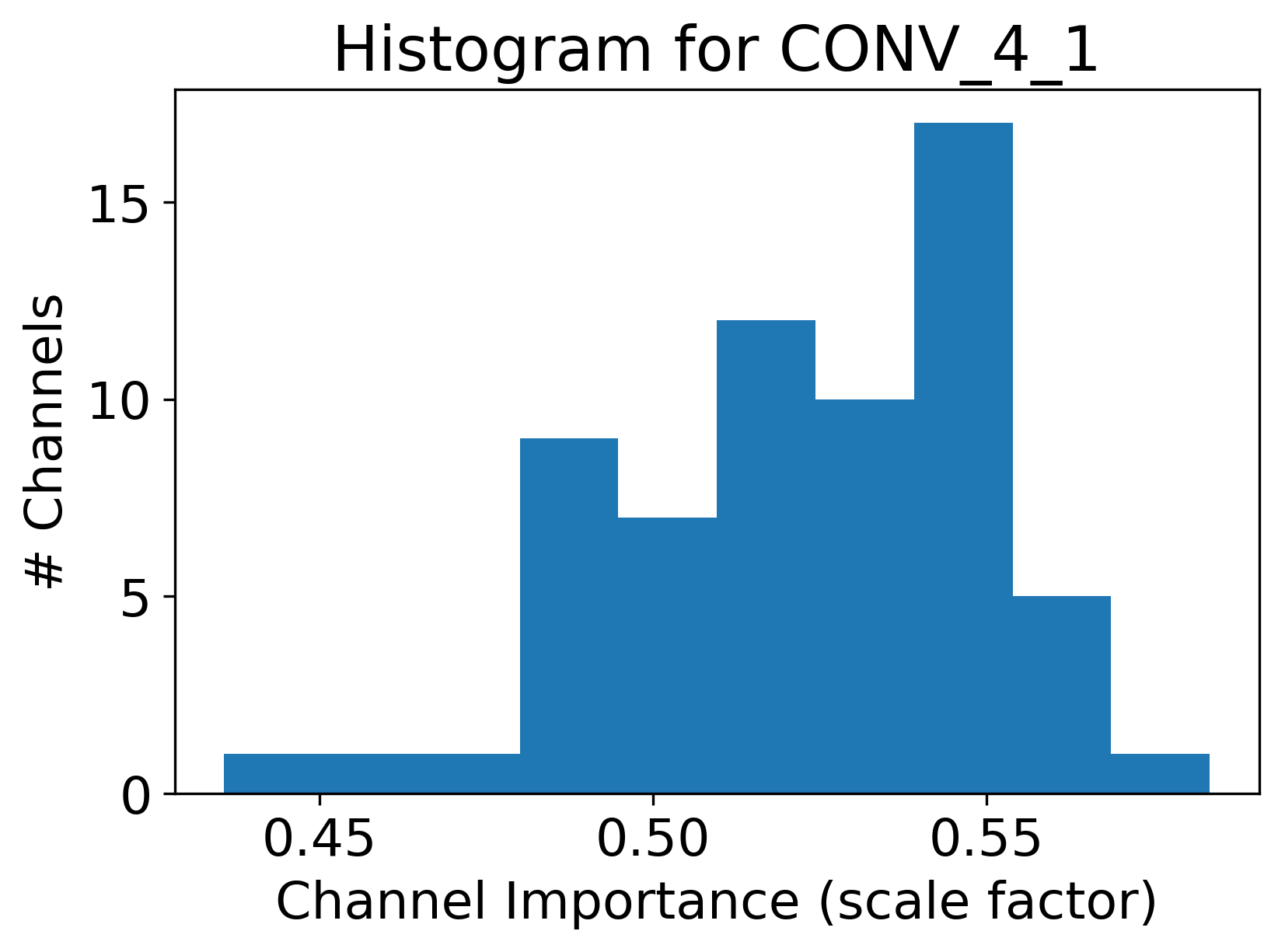}
   
    \caption{Figure shows the histogram of average channel-wise re-calibration weights (scale factor) for the Conv4\_1 layer of AB ResNet-56 trained on CIFAR-10}
    \label{fig:hist}
    \vskip -0.2in
   
\end{figure}

Fig \ref{fig:hist} shows a histogram over the channel (feature map) re-calibration weights produced by the AB block on the Conv4\_1 layer, i.e., the first layer of the last block, of AB ResNet-56 after being trained on CIFAR-10. The re-calibration weights were averaged over 10,000 test images in the CIFAR-10 dataset. We can see that the network gives different re-calibration weights to different channels. 

\begin{table}[t]
  \begin{center}
    \caption{Classification accuracy $\%$ on CIFAR-10 dataset for the ResNet-56 network with AB block by zeroing out the high value re-calibration weights and by zeroing out the low value re-calibration weights.}
    \label{tab:relevcalib}
    \scalebox{0.9}{
     \begin{tabular}{|c|c|c|} 
      \hline
     \textbf{$\%$ of Channels} & \textbf{Acc($\%$) High wts.} & \textbf{Acc($\%$) Low wts.} \\ 
      \hline
      $0$ (baseline AB) & $94.66$ & $94.66$ \\
      \hline
      $5$ &  $76.13$ & $91.32$ \\
      \hline
      $10$ & $53.52$ & $84.50$\\
     \hline
     $15$ & $40.01$ & $74.26$\\
     \hline
     $20$ & $29.24$ & $63.11$\\
     \hline
     $25$ & $18.22$ & $52.40$\\
     \hline
    \end{tabular}}
  \end{center}
\vskip -0.3in
\end{table}

The calibration weights produced by the AB block, re-calibrates the channels (feature maps) of the output produced the Convolutional filters. We perform two types of experiments to find out how these weights affect the network performance.  First, we progressively zero out x$\%$ of the highest channel re-calibration weights at each AB block. Second, we progressively zero out x$\%$ of the lowest channel re-calibration weights at each AB block. x is varied from 5$\%$ to 25$\%$.

Table \ref{tab:relevcalib} shows that, if we zero out the top $5\%$ high value channel re-calibration weights, the classification accuracy crashes to $76.13\%$ from $94.66\%$. This catastrophic drop continues as we increase the percentage of the high-value channel re-calibration weights that are to be zeroed, falling to $18.22\%$ after only $25\%$ high-value channel re-calibration weights have been zeroed. On the other hand, if we zero out the top $5\%$ low-value channel re-calibration weights, the classification accuracy drops by only $3\%$ and by the time we zero out $25\%$ low-value channel re-calibration weights, the classification accuracy is still at $52.4\%$, which is much higher than the other case. This signifies that those channels which had higher re-calibration weights had highly relevant features in them and zeroing them caused a drastic drop in the classification accuracy. Whereas, those channels which had lower re-calibration weights had not so relevant features and zeroing them could not affect the classification accuracy in such a drastic manner as the most relevant channels were still functioning. Therefore, we can conclude that the channels for which the AB block gives high re-calibration weights contain highly relevant features.

\section{Conclusion}
\label{conclusion}

In this paper, we proposed the Accuracy Booster block, a performance boosting block for CNNs that uses channel-wise (feature map) re-calibration. Our analysis showed how the AB block is lighter than other recent approaches. Through our several experiments, we show that the AB block performs consistently better than other designs of higher complexity. In our ablation study, we justified our architectural choices while experimenting on various architectures. Since the SE blocks were introduced, the general trend has been to further increase the complexity of such blocks to improve the performance of the base model. However, through extensive ablation studies, we show that too much increase in the complexity of such blocks may not always increase the model performance. We also show that our architecture generalizes to detection as well. Therefore, the Accuracy Booster block is a useful tool to be utilized in Neural Networks for boosting their performance.

\newpage
{\small
\bibliographystyle{ieee}
\bibliography{egbib}
}

\end{document}